\documentclass{article}

\usepackage[final,nonatbib]{neurips_2023}
\usepackage[utf8]{inputenc} % allow utf-8 input
\usepackage[T1]{fontenc}    % use 8-bit T1 fonts
\usepackage{hyperref}       % hyperlinks
\usepackage{url}            % simple URL typesetting
\usepackage{booktabs}       % professional-quality tables
\usepackage{amsfonts}       % blackboard math symbols
\usepackage{nicefrac}       % compact symbols for 1/2, etc.
\usepackage{microtype}      % microtypography
\usepackage{xcolor}         % colors
\usepackage{color, colortbl}

\definecolor{LightCyan}{rgb}{0.88,1,1} %#b9f2f0
\definecolor{Pink}{rgb}{1, 0.71, 0.756} %#ffb5c1
\definecolor{PaleGreen}{rgb}{0.8, 1, 0.8} %#ccffcc
\definecolor{PaleBlue}{rgb}{0.8, 0.9, 1} %#cce5ff
\definecolor{Orange}{RGB}{255, 180, 130} %#ffb482
\definecolor{Purple}{RGB}{208, 187, 255} %#d0bbff
\definecolor{BrightPink}{RGB}{250, 176, 228} %#fab0e4
\definecolor{Yellow}{RGB}{255, 254, 163} %#fffea3
\definecolor{Brown}{RGB}{222, 187, 155} %#debb9b

\usepackage{times}
\usepackage{epsfig}
\usepackage{graphicx}
\usepackage{amsmath}
\usepackage{amssymb}
\usepackage{diagbox}
\usepackage{adjustbox}

\usepackage[symbol]{footmisc}

\usepackage{tikz}
\usepackage{caption}
\usepackage{subcaption}

\usepackage{wrapfig}
\usepackage{comment}

\usepackage[textsize=tiny]{todonotes}
\usepackage{blindtext}
\usepackage{multirow}
\usepackage{makecell}

\definecolor{light-gray}{gray}{0.95}

\usepackage{graphicx}
\usepackage{array}
\usepackage{booktabs}
\newcolumntype{N}{>{\centering\arraybackslash}m{.5in}}
\newcolumntype{G}{>{\centering\arraybackslash}m{1in}}
\newcolumntype{H}{>{\centering\arraybackslash}m{1.1in}}
\newcolumntype{M}{>{\centering\arraybackslash}m{.3in}}
\newcolumntype{P}{>{\centering\arraybackslash}m{.8in}}
\newcolumntype{Q}{>{\centering\arraybackslash}m{.4in}}

\begin{document}

%%%%%%%%% TITLE
\title{Stable LM 2 1.6B Technical Report}

\author{
Marco Bellagente\textsuperscript{*} \quad Jonathan Tow\textsuperscript{*} \quad Dakota Mahan \quad Duy Phung \quad Maksym Zhuravinskyi \\
Reshinth Adithyan \quad James Baicoianu \quad Ben Brooks \quad Nathan Cooper \quad
Ashish Datta \\ 
Meng Lee \quad Emad Mostaque \quad Michael Pieler \quad
Nikhil Pinnaparju \quad Paulo Rocha \\  
Harry Saini \quad Hannah Teufel \quad Niccolo Zanichelli \quad Carlos Riquelme
\\
\\
\textbf{Stability AI Language Team}\thanks{Equal contribution. Correspondance to: \{marco.bellagente, jonathantow, carlos.riquelme\}@stability.ai}
}
\maketitle

%%%%%%%%% ABSTRACT

\begin{abstract}
% Intro
We introduce StableLM 2 1.6B, the first in a new generation of our language model series.
% The model
In this technical report, we present in detail the data and training procedure leading to the base and instruction-tuned versions of StableLM 2 1.6B.
The weights for both models are available via Hugging Face for anyone to download and use \footnote{\url{https://huggingface.co/stabilityai/stablelm-2-1_6b}}\footnote{\url{https://huggingface.co/stabilityai/stablelm-2-zephyr-1_6b}}.
% The details we share
The report contains thorough evaluations of these models, including zero- and few-shot benchmarks, multilingual benchmarks, and the MT benchmark focusing on multi-turn dialogues.
At the time of publishing this report, StableLM 2 1.6B was the state-of-the-art open model under 2B parameters by a significant margin.
% Other things
Given its appealing small size, we also provide throughput measurements on a number of edge devices.
In addition, we open source several quantized checkpoints and provide their performance metrics compared to the original model.
\end{abstract}

\section{Introduction}

Following the development of the Transformer architecture \cite{vaswani2017attention}, a remarkable number of proprietary and open-source large language models have been trained and deployed.
While countless new ideas and artifacts are announced weekly or daily, some key aspects remain opaque -- especially around the most powerful models.
Often, the training data is not disclosed.
This poses a fundamental challenge in times where society demands transparency as it faces a brand-new disruptive technology that is hard to inspect and audit.
In this report, we explain in a reproducible manner how to train a modest-size but state-of-the-art language model.
All the data we used is public (see Table \ref{tab:data-mix}) and its training required around 92k GPU hours of training -- worth around \$322k on popular cloud providers (assuming \$3.5 per GPU hour).
We hope this work contributes to the open AI community and helps set the standard for upcoming transparent models.

% \marco{filler ends here}\\
This report is organized as follows: Section~\ref{sec:pre-training} details the process of pre-training Stable LM 2 1.6B.  We devote Section~\ref{sec:alignment} to fine-tuning and human preference alignment.
Section~\ref{sec:experiments} presents model evaluations on standard downstream benchmarks.
Compiling and running inference Stable LM 2 on several edge devices is outlined in Section~\ref{sec:inference}.
We consider several follow-up directions in Section~\ref{sec:future}. Carbon emissions and societal impacts related to the training and release of Stable LM 2 are covered in Section~\ref{sec:impact}.
Finally, Section~\ref{sec:conclusion} concludes and summarizes this work.

\section{Model Pre-Training} \label{sec:pre-training}
    The first stage in training large language models (LLMs) focuses on learning to predict the next token in a sequence using a vast and diverse array of data sources. We refer to this stage as \emph{pre-training}.
    It enables models to build general-purpose internal representations suitable for basic language capabilities and even more advanced generation and comprehension tasks. In fact, it has been hypothesized that the majority of model knowledge and capabilities are learned during pre-training~\cite{zhou2023lima}. In this section, we introduce the design principles and ablations that influenced the creation of our training set, as well as details about the model architecture and training procedure. While many similar reports exist for other cutting-edge models, they often omit critical details, such as the particular data sources, sampling weights, or the complete set of ablations they performed.
    As a result, the open-source community cannot accurately reproduce these models.
    On the other hand, we present a fully transparent log of our model training details. We are confident that researchers and practitioners will find valuable insights in this comprehensive account.
\begin{table}[ht]
\centering
\noindent
\begin{tabular}{G N G N N N}
\toprule
%\begin{tabular}{N N N N N }\toprule
\multicolumn{1}{ c }{\textbf{Dataset}} & \textbf{Sampling Weight} & \textbf{Num Tokens} & \textbf{Epochs} & \textbf{Category}\\[1ex] 
\cmidrule(lr){1-1}
\cmidrule(lr){2-5}
\rowcolor{PaleBlue}
\multicolumn{1}{ l }{\textbf{Arxiv}} & 0.0079 & 15,852,446,656 &0.75& Academic\\ [0.2ex]
\rowcolor{PaleBlue}
\multicolumn{1}{ l }{\textbf{PubMed}} & 0.012 &24,126,600,626 &1.0& Academic\\ [0.2ex] 
\rowcolor{PaleBlue}
\multicolumn{1}{ l }{\textbf{S2ORC}} & 0.0318 &63,910,567,858 &1.0& Academic\\ [0.2ex]
\rowcolor{PaleBlue}
\multicolumn{1}{ l }{\textbf{PhilPapers}} & 0.0013 &2,591,115,796 &4.0& Academic\\ [0.2ex]
\cmidrule(lr){1-1}
\cmidrule(lr){2-5}
\rowcolor{Orange}
\multicolumn{1}{ l }{\textbf{BookCorpusOpen}} & 0.0045 &9,135,954,810 &6.0& Books\\ [0.2ex]
\rowcolor{Orange}
\multicolumn{1}{ l }{\textbf{PG-19}} & 0.0091 &18,293,226,468 &4.0& Books\\ [0.2ex]
\rowcolor{Orange}
\multicolumn{1}{ l }{\textbf{FanFics}} & 0.0018 &3,644,583,700 &4.0& Books\\ [0.2ex]
\cmidrule(lr){1-1}
\cmidrule(lr){2-5}
\rowcolor{PaleGreen}
\multicolumn{1}{ l }{\textbf{Cultura-X (EN)}} & 0.2521 &506,625,211,049&0.72& Web\\ [0.2ex] 
\rowcolor{PaleGreen}
\multicolumn{1}{ l }{\textbf{Cultura-X (ES)}} & 0.0155 &31,241,701,156&0.4& Web\\ [0.2ex] 
\rowcolor{PaleGreen}
\multicolumn{1}{ l }{\textbf{Cultura-X (DE)}} & 0.0152 &30,628,813,934&0.32& Web\\ [0.2ex] 
\rowcolor{PaleGreen}
\multicolumn{1}{ l }{\textbf{Cultura-X (FR)}} & 0.0097&19,466,611,808 &0.26& Web\\ [0.2ex] 
\rowcolor{PaleGreen}
\multicolumn{1}{ l }{\textbf{Cultura-X (IT)}} & 0.0096&19,202,903,491 &0.4& Web\\ [0.2ex] 
\rowcolor{PaleGreen}
\multicolumn{1}{ l }{\textbf{Cultura-X (NL)}} & 0.0097& 19,511,322,386  &0.62& Web\\ [0.2ex] 
\rowcolor{PaleGreen}
\multicolumn{1}{ l }{\textbf{Cultura-X (PT)}} & 0.01& 20,063,968,694  &0.78& Web\\ [0.2ex] 
\rowcolor{PaleGreen}
\multicolumn{1}{ l }{\textbf{C4}} & 0.0855 & 171,782,178,108&1.0& Web\\ [0.2ex] 
\rowcolor{PaleGreen}
\multicolumn{1}{ l }{\textbf{OpenWebText2}} & 0.0130 & 26,161,864,434  &3.0& Web\\ [0.2ex]
\rowcolor{PaleGreen}
\multicolumn{1}{ l }{\textbf{RefinedWeb}} & 0.3292 & 661,591,178,339 &1.15& Web\\ [0.2ex] 
\cmidrule(lr){1-1}
\cmidrule(lr){2-5}
\rowcolor{Purple}
\multicolumn{1}{ l }{\textbf{StackExchange}} &0.0231&46,302,993,820 &2.5& Social\\ [0.2ex]
\rowcolor{Purple}
\multicolumn{1}{ l }{\textbf{HackerNews}} &0.0019&3,817,060,582 &2.0& Social\\ [0.2ex] 
\cmidrule(lr){1-1}
\cmidrule(lr){2-5}
\rowcolor{Pink}
\multicolumn{1}{ l }{\textbf{EuroParl}} & 0.0023&4,678,506,882 &3.0& Law\\ [0.2ex] 
\rowcolor{Pink}
\multicolumn{1}{ l }{\textbf{FreeLaw}} & 0.0088 &17,694,697,577 &1.2& Law\\ [0.2ex] 
\rowcolor{Pink}
\multicolumn{1}{ l }{\textbf{PileOfLaw}} & 0.0063&12,663,061,642 &0.75& Law\\ [0.2ex] 
\cmidrule(lr){1-1}
\cmidrule(lr){2-5}
\rowcolor{LightCyan}
\multicolumn{1}{ l }{\textbf{DM Math}} & 0.0066 &13,321,872,138 &3.0& Math\\ [0.2ex] 
\rowcolor{LightCyan}
\multicolumn{1}{ l }{\textbf{AMPS}} & 0.0011 &2,126,719,278 &6.0& Math\\ [0.2ex] 
\rowcolor{LightCyan}
\multicolumn{1}{ l }{\textbf{OpenWebMath}} & 0.0132 &26,530,341,292 &2.0& Math\\ [0.2ex] 
\cmidrule(lr){1-1}
\cmidrule(lr){2-5}
\rowcolor{BrightPink}
\multicolumn{1}{ l }{\textbf{RedPajama Wiki}} & 0.0363 &72,866,870,472 &3.0& Wiki\\ [0.2ex] 
\cmidrule(lr){1-1}
\cmidrule(lr){2-5}
\rowcolor{Yellow}
\multicolumn{1}{ l }{\textbf{Starcoder}} & 0.0724 & 145,586,775,301&0.74& Code\\ [0.2ex] 
\cmidrule(lr){1-1}
\cmidrule(lr){2-5}
\rowcolor{Brown}
\multicolumn{1}{ l }{\textbf{Restruct-v1}} & 0.0102 &20,412,655,632 &3.0& Instruction\\ [0.2ex] 
\midrule
\multicolumn{1}{ l }{\textbf{Total}} & 1 &2,009,831,803,933 && -\\ [0.2ex] 
\end{tabular}
\caption{The complete Stable LM 2 training set with sampling weights. The tokens count refers to the \textbf{Arcade100k} tokenizer introduced in Sec.~\ref{sec:tokenizer}.
The number of tokens in the table already includes the number of epochs shown next to it.
For instance, in the case of BookCorpusOpen, we use around 9B tokens, corresponding to 6 epochs of the original dataset (that is, each epoch is around 1.5B tokens).
Similarly, if epochs are below one, the number of tokens shown is a subset of the total dataset.
}
\label{tab:data-mix}
\end{table}
\begin{figure}
    \centering
    \includegraphics[width=0.95\linewidth]{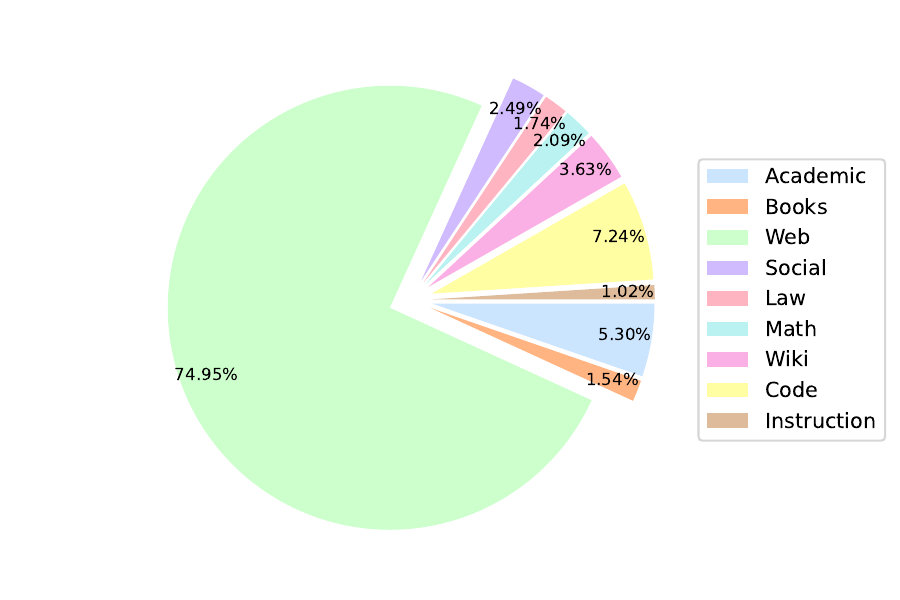}
    \caption{Percentage of effective training tokens by domain in the Stable LM 2 pre-training dataset.}
    \label{fig:data_by_source}
\end{figure}

\subsection{Training}
We train Stable LM 2 to predict the next token following standard autoregressive sequence modeling~\cite{radfordimproving}. We
train our model from scratch with a context length of 4096 and benefit from the efficient sequence-wise parallelism optimizations of FlashAttention-2~\cite{dao2022flashattention, dao2023flashattention2}.
Training is performed in BFloat16 mixed precision while keeping all-reduce operations in FP32. \cite{chowdhery2022palm, wortsman2023smallscale} found it beneficial to add a z-loss regularization term on the softmax normalizer, $z_{\mathrm{loss}} \propto \log^2 Z$, to mitigate instability stemming from output logits divergence.
While it did not hurt performance in our ablations, the improvements to stability were minimal.
Accordingly, it was not applied in the final run.
We employ a standard AdamW optimizer with the following hyperparameters: $\beta_1 = 0.9, \beta_2 = 0.95, \epsilon = 1e-8,  \lambda (\text{weight decay}) = 0.1$.
Sec.~\ref{sec:scheduler} offers details regarding the custom learning rate scheduler that we applied.

\subsection{Data}
Model performance is affected by pre-training data design decisions, including both the source selection and the sampling weights~\cite{longpre2023pretrainers}. Our approach is close to that of~\cite{touvron2023llama}: the majority of our training data consists of sources utilized in the training of other LLMs, such as RefinedWeb~\cite{penedo2023refinedweb}, subsets of the Pile~\cite{gao2020pile}, RedPajama~\cite{together2023redpajama} and the Stack~\cite{li2023starcoder}. We supplement these with OpenWebText~\cite{Gokaslan2019OpenWeb}, OpenWebMath~\cite{paster2023openwebmath}, and parts of CulturaX~\cite{nguyen2023culturax}. While inspecting randomly sampled documents from the mC4 subset of CulturaX, we encountered frequent HTML boilerplate and decided to remove this portion entirely, finally keeping only the OSCAR subset.
Additionally, we incorporate FanFics\footnote{\url{https://huggingface.co/datasets/atom-in-the-universe/fanfics-10k-50k}}, a subset of 50k documents from \emph{fanfiction.net} selected by lowest perplexity scores according to a KenLM\footnote{\url{https://huggingface.co/edugp/kenlm}}. Finally, following~\cite{yuan2022restructured}, we restructure several raw datasets into rich fixed forms for downstream tasks such as summarization, question-answering, sentiment analysis, etc., and add instruction data from \cite{longpre2023data}, the aggregate of which we collectively call Restruct-v1. The list of sources used in Restruct-v1 is made available in Tab.~\ref{tab:restruct-data}.
Stable LM's training set is comprised entirely of open-source datasets compatible with commercial usage, most of which are hosted on the Hugging Face Hub.
The only exception to the latter aspect (HFH), Restruct-v1, can easily be reproduced by following the approaches and prompt templates provided by \cite{yuan2022restructured}.

Carefully selecting the mixture proportions of the various data domains is critical, particularly with respect to the amount of non-English and code data.
We trained several models on different mixes and evaluated them on downstream benchmarks to pick our final dataset.
The full set of ablations is available in Appendix~\ref{sec:data-ablations}, together with rationales for selecting these particular mixes. 
Based on the results of the ablations, we trained our model with the mix shown in Table~\ref{tab:data-mix}, which accounts for around 2 trillion tokens.
Note that it includes multilingual data in German (DE), Spanish (ES), French (FR), Italian (IT), Dutch (NL), and Portuguese (PT). The split of our dataset across different domains is visualized in Fig.~\ref{fig:data_by_source}.
\begin{table}
\centering
\noindent
\begin{tabular}{P P P P G}\toprule
\textbf{Parameters} & \textbf{Hidden Size} & \textbf{Layers} & \textbf{Heads} & \textbf{Sequence Length} \\[0.1ex]
\midrule
1,644,417,024 & 2048 & 24 & 32 & 4096 \\ [1ex]
\midrule
\end{tabular}
\caption{Stable LM 2 model architecture.}
\label{tab:architecture-layout}
\begin{tabular}{H H H H}
\toprule
\textbf{Data Parallel Degree} & \textbf{Micro Batch Size} & \textbf{Gradient Accumulation Steps} & \textbf{Activation Checkpointing} \\[0.1ex]
\midrule
512 & 2 & 2 & disabled \\ [0.1ex]
\midrule
\end{tabular}
\caption{Stable LM 2 training configuration.}
\label{tab:training-config}
\end{table}

% High level description of the design choices for the training set selection
% \begin{itemize}
%     \item Up-sample high-quality sources over web data
%     \item Maximize multilingual performance without sacrificing English
% \end{itemize}

\subsection{Tokenizer}\label{sec:tokenizer}
We use \textbf{Arcade100k}, a BPE tokenizer extended from OpenAI's \verb|tiktoken.cl100k_base| to include special tokens for code~\cite{li:starcoder} and digit-split handling~\cite{liu2023goat,bai2023qwen}. The vocabulary consists of 100,289 tokens and is padded to the nearest multiple of 64 (100,352) during training to meet the recommended Tensor Cores alignment on NVIDIA A100 devices. In preliminary experiments, we did not observe statistically significant deviations in downstream natural language performance tasks when compared against a hyperparameter matching model trained with the smaller GPT-NeoX tokenizer~\cite{black2022gptneox20b}. Increased compression rates for code and non-English languages influenced our decision to use \textbf{Arcade100k} over existing tokenizers.
\subsection{Architecture and Training Layout}
The model is a causal decoder-only transformer similar in design to the LLaMA architecture~\cite{touvron2023llama}. Table~\ref{tab:architecture-layout} shows some of the key architectural details.
In particular, the main differences with respect to LLaMA are the following:
\begin{itemize}
    \item 
    \textbf{Position Embeddings}. Rotary Position Embeddings~\cite{su2023roformer} applied to the first $25\%$ of head embedding dimensions for improved throughput following~\cite{black2022gptneox20b}.
    \item 
    \textbf{Normalization}. LayerNorm~\cite{ba2016layer} with learned bias terms as opposed to RMSNorm~\cite{zhang2019root}.
    \item 
    \textbf{Biases}. We remove all bias terms from the feed-forward networks and multi-head self-attention layers, except for the biases of the key, query, and value projections~\cite{bai2023qwen}.
\end{itemize}

Stable LM 2 1.6B is trained on 64 Amazon P4d instances comprising 512 NVIDIA A100 (40GB HBM2) GPUs. The size of our model, together with ZeRO stage 1 distributed optimization~\cite{rajbhandari2020zero}, eliminates the need for model sharding. Still, different triplets of \textbf{micro batch size}, \textbf{gradient accumulation steps}, and \textbf{activation checkpointing} granularity lead to different speed metrics.
Following the recommendations in~\cite{hagemann2023efficient}, we obtain our final configuration by finding the micro-batch size that allows us to completely remove activation checkpointing. We then determine the gradient accumulation steps based on our target batch size and \textbf{data parallel} degree. We employ a batch size of $8,388,608$ tokens, based on the observations in Appendix~\ref{sec:batch}. With the setup in Table~\ref{tab:training-config}, we achieve $\approx$170 TFLOPs/s per device, or $54.5\%$ model flops utilization (MFU). A higher hardware utilization of $\approx$200 TFLOPs/s ($64\%$ MFU) can be trivially achieved by decreasing the degree of data parallelism and correspondingly increasing the number of gradient accumulation steps, at the cost of an increased iteration time.
%
% %
% \begin{table}
%     \centering
%     \begin{tabular}{c c c c c}
%      Parameters	& Hidden Size &	Layers &	Heads &	Sequence Length \\[0.1ex]
%      \cmidrule{1-5}
%      1,644,417,024 & 2048 & 24 & 32 & 4096
% \end{tabular}
% \caption{Stable lm 2 model architecture.}
% \label{tab:architecture}
% \end{table}
% %
%
% \begin{table}
% \centering
% \begin{tabular}{cccc}
%      dp & mbs & gas & activation checkpointing\\[0.1ex]
%      \cmidrule{1-4}
%      512 & 2 & 2 & disabled\\
% \end{tabular}
% \caption{Layout configuration used for pre-training stable lm 2.}
% \label{tab:layout}
% \end{table}
%
\subsection{Learning Rate Scheduler}\label{sec:scheduler}
We propose a new learning rate scheduler.
It consists of multiple stages and is designed to favor flexibility in terms of continued pre-training.
We begin by linearly increasing the learning rate to its max value of $1e-3$ over 9720 steps. This \emph{warm up} stage is followed by the main training phase in which the learning rate is decreased according to Eq.~\ref{eq:scheduler}:
\begin{equation}\label{eq:scheduler}
  \left\{
    \begin{aligned}
      & m + \frac{(M - m)}{2} * \left[ \cos \left( 2\pi * \frac{i}{N} \right)  + 1 \right] \quad &\text{if} \quad i \leq N/4 &\qquad \text{\emph{cosine} decay} \\
      & \frac{\alpha}{\sqrt{i + \beta}} \quad &\text{if} \quad i > N/4 &\qquad \text{\emph{rsqrt} decay}
    \end{aligned}
  \right.
\end{equation}
where $m$ and $M$ are respectively the minimum and maximum learning rates, $i$ is the current step, and $N$ is the total number of steps. The free parameters $\alpha$ and $\beta$ have been arbitrarily chosen to enforce the continuity of the scheduler and its derivative at $i=N/4$. 
We finalize training by linearly cooling the learning rate down to zero over 80k steps, which corresponds to around 670B tokens.
%In this phase, we also increase the sampling rate of restructured pre-training data from 1\% to 4\% while downsampling multilingual web subsets.%
%\textcolor{red}{TODO: @Jon fill this in better - can we add plot? table? }
%
The full scheduler is illustrated in Fig.~\ref{fig:hybrid-scheduler}; further details and ablations can be found in Appendix~\ref{app:scheduler}.

\begin{figure}
    \centering
    \includegraphics[width=8cm]{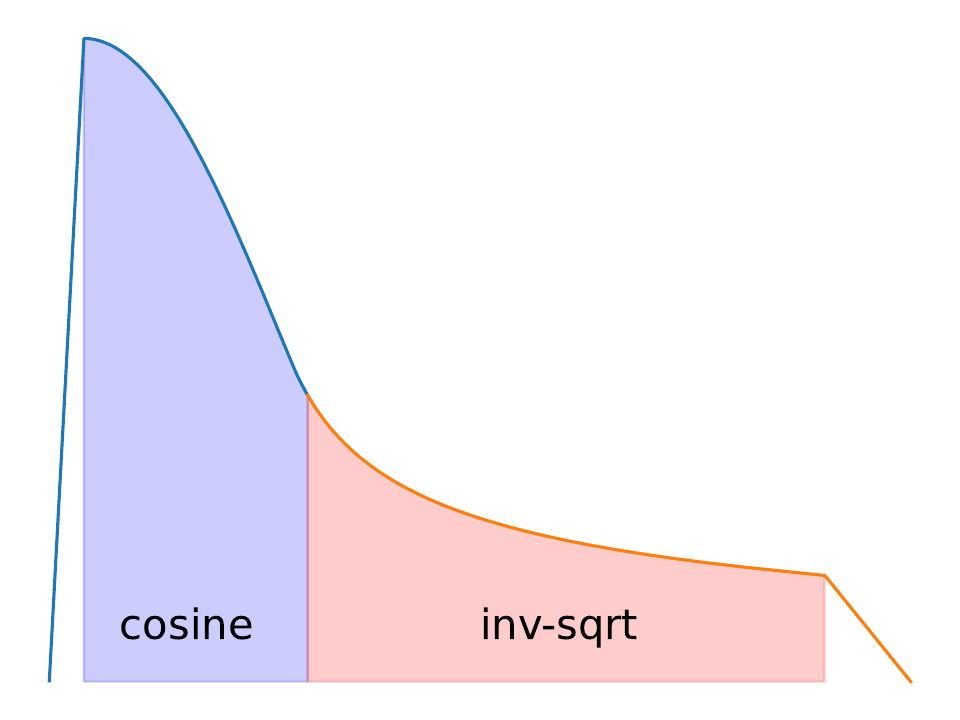}
    \caption{Multi-stage infinite scheduler proposed and applied in this work.}
    \label{fig:hybrid-scheduler}
\end{figure}
\section{Fine-tuning and Alignment} \label{sec:alignment}

% \carlos{Add here brief high level intro regarding what we do and cover in this section. I added a basic one. Max and Duy please check my multilinguality statement.}

Following pre-training, we further develop the conversational skills of our model via a fine-tuning stage that consists of three main steps: supervised fine-tuning (SFT), direct preference optimization (DPO), and self-knowledge learning.
Importantly, we do not use multilingual data at this stage.
We now describe each of them in detail, and in Section \ref{sec:experiments} we report the results after all three steps.
% \hannah{Might be nice to have tables similar to Table 1 for the SFT and DPO datasets - especially if we want to share as much as possible for the stanford report} %% Good point, added.

\begin{table}
\centering
\noindent
\begin{tabular}{M M M M M M N M M}\toprule
%\begin{tabular}{N N N N N }\toprule
\multicolumn{1}{ l }{\textbf{Model}} & \textbf{Size} & \textbf{Avg} & \textbf{ARC} & \textbf{HS} & \textbf{MMLU} & 
\textbf{TQA} & \textbf{Wino} & \textbf{GSM} \\[0.1ex]
\cmidrule(lr){1-1}
\cmidrule(lr){2-9}
\multicolumn{1}{ l }{\textbf{phi-2}\dag} & 2.8B & 61.3&	61.1 &	75.1 &	58.1&	44.5 &	74.4&	54.8 \\ [0.1ex]
\rowcolor{PaleBlue}\multicolumn{1}{ l }{\textbf{stablelm-2-zephyr-1\_6b}\dag} & 1.6B & 49.7&	43.3&	69.3&	41.8&	45.6&	63.6&	34.8\\ [0.1ex]
\multicolumn{1}{ l }{\textbf{phi-1\_5}\dag} & 1.3B & 47.7& 52.9& 63.8& 43.9& 40.9& 72.2& 12.4 \\ [0.1ex]
\multicolumn{1}{ l }{\textbf{stablelm-3b-4e1t}} & 2.7B & 46.6 & 46.6 & 75.9 & 45.2 & 37.2 & 71.2 & 3.3 \\ [0.1ex]
\multicolumn{1}{ l }{\textbf{Qwen-1.5-1.8B}} & 1.8B & 46.6 & 37.9 & 61.4 & 46.7 & 39.4 & 60.3 & 33.6\\ [0.1ex]
\multicolumn{1}{ l }{\textbf{gemma-2b}} & 2.5B & 46.5 & 48.5 & 71.7 & 41.7 & 33.1 &66.8 & 17.4\\ [0.1ex]
\rowcolor{PaleBlue} \multicolumn{1}{ l  }{\textbf{stablelm-2-1\_6b}} & 1.6B & 45.3 & 43.3 & 70.5 & 38.9 & 36.8 & 64.6 & 17.4 \\ [0.1ex]

\multicolumn{1}{ l }{\textbf{gemma-2b-it}\dag} & 2.5B & 42.7 & 43.9 & 62.7 & 37.6 & 45.8 & 60.9 & 5.5\\ [0.1ex]

\multicolumn{1}{ l }{\textbf{open\_llama\_3b}} & 3B & 40.3 & 39.9 & 71.6 & 27.1& 34.8 & 67.0 & 0.9 \\ [0.1ex]
%\multicolumn{1}{ l }{\textbf{h2o-danube-1.8b}} &1.8B & 39.1 & 39.4 & 69.6 & 25.9& 33.9& 64.5& 1.4 \\ [0.1ex]
\multicolumn{1}{ l }{\textbf{falcon-rw-1b}} &1.3B & 37.1 & 35.1 & 63.6 & 25.3 & 35.9 & 62.0 & 0.5 \\ [0.1ex]
%\multicolumn{1}{ l }{\textbf{opt-2.7b}} &2.7B & 36.7 & 33.9 & 61.4 &25.4& 37.4 & 61.9 & 0.2 \\ [0.1ex]
\multicolumn{1}{ l }{\textbf{TinyLLama-1.1B}} & 1.1B & 36.4 & 33.9 & 60.3 &26.0 &37.3 &59.5 &1.4 \\ [0.1ex]
\midrule
\end{tabular}
\caption{Comparison of Open LLM leaderboard evals. \dag \ denotes aligned models. Note that in this table, as well as in Tab.~\ref{tab:pre-training-eval-multilingual}, we marked the Phi series of models~\cite{gunasekar2023textbooks} as \emph{aligned}. While we acknowledge that they may have only been pre-trained, by the nature of the training data used, and the self-disclaimed intended use for QA and chat formats, we believe this to be a fair statement.}
\label{tab:pre-training-eval}
\end{table}
\begin{table}
\centering
\noindent
\begin{tabular}{Q M M M M M M M M}\toprule
\multicolumn{1}{ l }{\textbf{Model}} & \textbf{All}& \textbf{EN} & \textbf{DE} & \textbf{ES} & \textbf{FR} & \textbf{IT} & \textbf{NL} & \textbf{PT}\\[0.1ex]
\cmidrule(lr){1-1}
\cmidrule(lr){2-9}
\multicolumn{1}{ l }{\textbf{stablelm-3b-4e1t}}& 41.7&  50.9&	39.7&	40.2&	41.2&	41.1&	36.7&	41.9 \\ [0.1ex]
\rowcolor{PaleBlue} \multicolumn{1}{ l }{\textbf{stablelm-2-zephyr-1\_6b}\dag} &41.5&  49.5&	40.2&	40.0&	39.8&	39.9&	38.8&	42.0\\ [0.1ex]
\rowcolor{PaleBlue} \multicolumn{1}{ l }{\textbf{stablelm-2-1\_6b}} &40.5&  48.7&	39.1&	39.0&	39.3&	38.8&	37.8&	41.2 \\ [0.1ex]
\multicolumn{1}{ l }{\textbf{gemma-2b}}& 39.8&  48.6&	38.3&	38.7&	38.7&	38.4&	35.1&	40.5\\ [0.1ex]
\multicolumn{1}{ l }{\textbf{gemma-2b-it}\dag}& 38.2&  49.4&	36.8&	38.0&	37.5&	35.5&	32.1&	38.1\\ [0.1ex]
\multicolumn{1}{ l }{\textbf{open\_llama\_3b}}& 37.5&  47.3&	35.2&	36.4&	37.6&	37.1&	32.2&	36.8\\ [0.1ex]
\multicolumn{1}{ l }{\textbf{Qwen-1.5-1.8B-Chat}}& 35.5&	46.2&	33.2&	35.1&	34.3&	33.2&	30.9& 35.7\\ [0.1ex]
\multicolumn{1}{ l }{\textbf{Qwen-1.5-1.8B}}& 34.8&  46.3&    31.8&   34.0&   34.2&    32.8&   29.7&   35.0\\ [0.1ex]
\multicolumn{1}{ l }{\textbf{TinyLLama-1.1B}}& 34.8&  42.4&	33.0&	33.8&	34.7&	33.5&	31.0&	35.0\\ [0.1ex]
\multicolumn{1}{ l }{\textbf{phi-2}\dag} & 34.6&  55.8&	29.0&	34.3&	32.9&	29.9&	27.1&	33.4\\ [0.1ex]
\multicolumn{1}{ l }{\textbf{falcon-rw-1b}}& 29.9&  42.2&	27.4&	28.6&	28.3&	28.0&	25.9&	29.1\\ [0.1ex]
\multicolumn{1}{ l }{\textbf{phi-1\_5}\dag}& 29.7&  47.1&	25.3&	28.7&	26.8&	27.2&	24.1&	29.0\\ [0.1ex]
\midrule
\end{tabular}
\caption{Multilingual evaluations. As in the previous table, \dag \ denotes aligned models.
}
\label{tab:pre-training-eval-multilingual}
\end{table}

\subsection{SFT}
The first step is supervised fine-tuning.
We fine-tune the pre-trained model on a number of instruction datasets publicly available on the Hugging Face Hub.
In particular, we use the following \emph{conversational} datasets: UltraChat \cite{ding2023enhancing}, WizardLM \cite{xu2023wizardlm}, SlimOrca \cite{SlimOrca}, ShareGPT \cite{wang2023openchat}, Capybara \cite{daniele2023amplify-instruct}, Deita \cite{liu2023what}, and MetaMathQA \cite{yu2023metamath}.
We removed any samples that exceeded eight turns, leading to a total of 826,938 samples.

% \carlos{Idea: provide a table --maybe in the appendix-- with one example per dataset?}

\begin{table}[t]
\centering
\begin{tabular}{l l l r}
\toprule
\textbf{Dataset} & \textbf{Type} & \textbf{Source} & \textbf{Number of Samples} \\[1ex]
\midrule
UltraChat & SFT & HuggingFaceH4/ultrachat\_200k & 194,409 \\ [0.2ex]
WizardLM & SFT & WizardLM/WizardLM\_evol\_instruct\_V2\_196k & 80,662 \\ [0.2ex]
SlimOrca & SFT & Open-Orca/SlimOrca-Dedup & 143,789 \\ [0.2ex]
ShareGPT & SFT & openchat/openchat\_sharegpt4\_dataset & 3,509 \\ [0.2ex]
Capybara & SFT & LDJnr/Capybara & 7,291 \\ [0.2ex]
Deita & SFT & hkust-nlp/deita-10k-v0 & 2,860 \\ [0.2ex]
MetaMathQA & SFT & meta-math/MetaMathQA & 394,418 \\ [0.2ex]
\midrule
Ultrafeedback & DPO & argilla/ultrafeedback-binarized-preferences & 63,169 \\ [0.2ex]
Orca Pairs & DPO & Intel/orca\_dpo\_pairs & 12,859 \\ [0.2ex]
\midrule
\end{tabular}
\caption{Fine-tuning datasets}
\label{tab:finetune-datasets}
\end{table}

%%% Carlos: commented this as links may not be needed?
% \subsubsection{Data}
% We collected high-quality of dialogue instruction datasets: \textbf{UltraChat}\footnote{\url{https://huggingface.co/datasets/HuggingFaceH4/ultrachat_200k}},\textbf{WizardLM}\footnote{\url{https://huggingface.co/datasets/WizardLM/WizardLM_evol_instruct_V2_196k}},\textbf{SlimOrca}\footnote{\url{https://huggingface.co/datasets/Open-Orca/SlimOrca}},\textbf{ShareGPT}\footnote{\url{https://huggingface.co/datasets/openchat/openchat_sharegpt4_dataset}},\textbf{Capybara}\footnote{\url{https://huggingface.co/datasets/LDJnr/Capybara}},\textbf{Deita}\footnote{\url{https://huggingface.co/datasets/hkust-nlp/deita-10k-v0}},\textbf{MetaMathQA}\footnote{\url{https://huggingface.co/datasets/meta-math/MetaMathQA}}, Subsequently, we eliminated any samples exceeding eight turns, resulting in the compilation of datasets comprising 888,423 samples.

% \subsubsection{Details of SFT training}
We train our SFT models for three epochs using a cosine learning rate scheduler. A warm-up phase of $10\%$ of the training duration is employed before reaching the peak learning rate of $5e-5$. We set the global batch size to 512 sequences and pack inputs into sequences of up to 4096 tokens in length.

\subsection{DPO}
Direct Preference Optimization (DPO)~\cite{radfordimproving} has been a fundamental tool in recent strong models such as Zephyr-7B \cite{tunstall2023zephyr}, Neural-Chat-7B, and Tulu-2-DPO-70B \cite{ivison2023camels}.
Accordingly, after applying SFT, we aligned the resulting model via DPO. We use two datasets at this stage: UltraFeedback ~\cite{cui2023ultrafeedback} and Intel Orca Pairs. We filter the datasets by removing pairs with tied ranking, pairs with duplicated content, and pairs in which the score for the chosen response is less than eight out of ten. We train the model with DPO following the Zephyr recipe~\cite{tunstall2023zephyr} and borrowing most of its hyperparameters, except for $\beta$, which we lower to $0.01$, and the learning rate, which we lower to $4e-6$, both of which helped improve the stability of training and final performance. This stage of training is performed using the Alignment Handbook~\cite{alignment_handbook2023}.

\begin{figure}
    \centering
    \includegraphics[width=12cm]{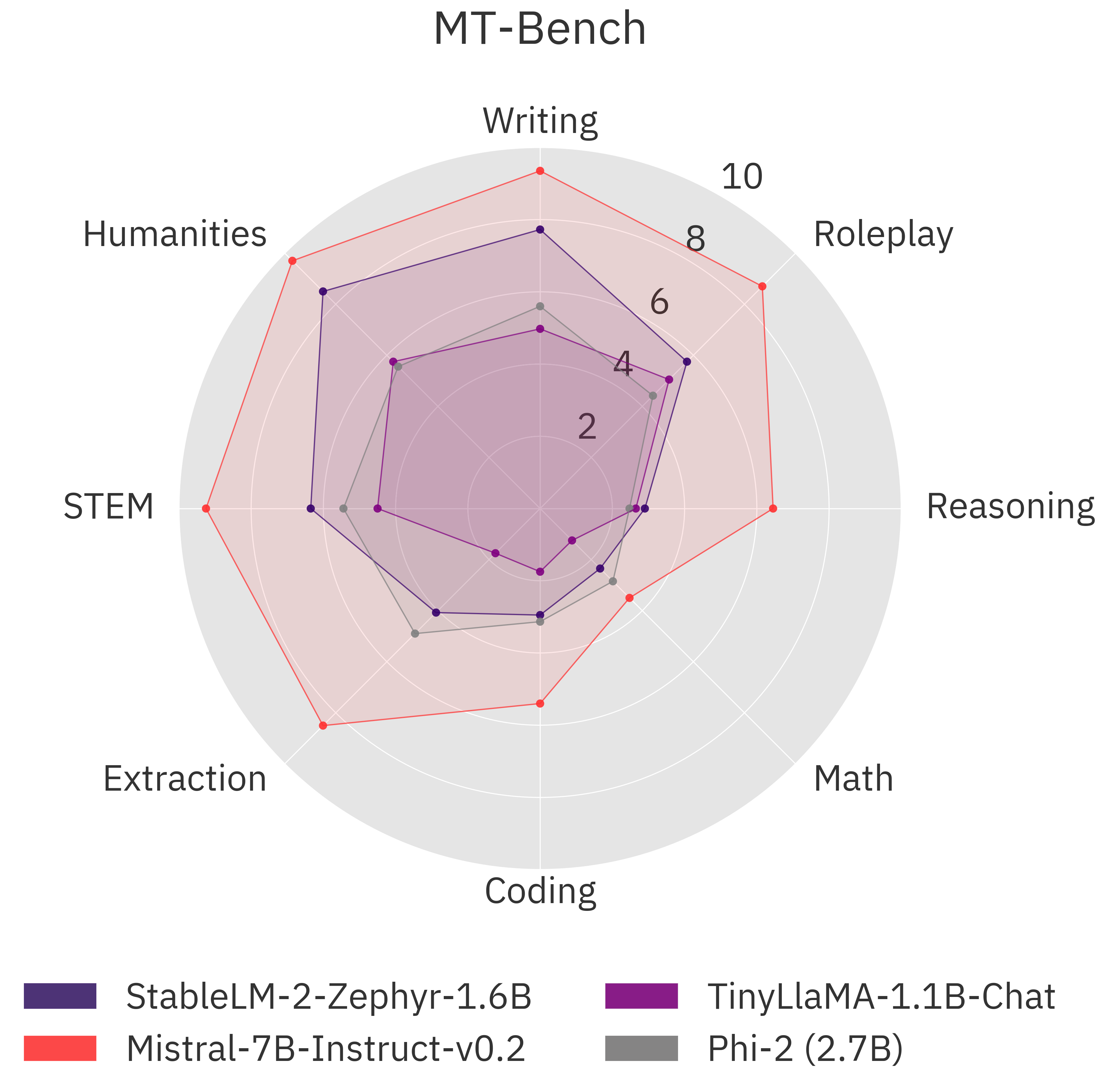}
    \caption{Stable LM 2 1.6B shows competitive performance, matching or even surpassing significantly larger models on MT-Bench.
    }
    \label{fig:MT-bench}
\end{figure}
%

% \textbf{UltraFeedback}\footnote{\url{https://huggingface.co/datasets/HuggingFaceH4/ultrafeedback_binarized}} and \textbf{Intel Orca}\footnote{\url{https://huggingface.co/datasets/Intel/orca_dpo_pairs}} datasets.

\subsection{Self-Knowledge}
The model after the output of the DPO~\cite{radfordimproving} stage does not have knowledge about who created it, or even what limitations a language model has. To remedy this, we were inspired by the data generation method of Reinforcement Learning from Contrast Distillation (RLCD)~\cite{yang2023rlcd} and the training method of Conditioned Reinforcement Learning Fine-Tuning (C-RLFT)~\cite{wang2023openchat}, which we apply to self-knowledge training.

To generate the initial prompts, we use the base model to generate 10k random first messages to a language model with no duplicates. To generate continuations, we use a few-shot prompt with self-knowledge in the previous chat turns as a positive completion. For the negative prompt, we simply sample from the prompt with no additional prompting or few-shot turns.

We train with unpacked examples for 6 epochs using a batch size of 256, a warmup stage for 100 steps to a maximum LR of 3e-6 followed by a cosine decay to zero. The positive prompts are trained in the same way as the SFT step, while the negative prompts are trained with a negative token instead of the Assistant token.

\section{Experimental Results and Benchmarks}
\label{sec:experiments}
This section presents the main experimental results for Stable LM 2 1.6B.
We compare with similarly-sized open-source models showing strong improvements on various tasks, including multilingual capabilities in Spanish, German, French, Italian, Portuguese, and Dutch.
For context, we also provide comparisons with much larger models.
We split our experiments into three sets: few- and zero-shot evaluations in English (as commonly done in the Hugging Face Open LLM leaderboard), multilingual evaluations, and conversational evaluations.

\subsection{Few-shot and Zero-shot Evaluations}

First, we assess the 0-shot and few-shot capabilities of Stable LM 2 by evaluating our model over popular benchmarks and comparing results against similarly sized open-source pre-trained models. 
Table~\ref{tab:pre-training-eval} presents model evaluations in English.
Our results cover the 6 benchmarks from the Open LLM Leaderboard (\cite{open-llm-leaderboard}): ARC-Challenge 25-shot~\cite{clark2018think} (\textbf{ARC}), HellaSwag 10-shot~\cite{zellers-etal-2019-hellaswag} (\textbf{HS}), MMLU 5-shot~\cite{hendrycks2021measuring} (\textbf{MMLU}), TruthfulQA 0-shot~\cite{lin2022truthfulqa} (\textbf{TQA}), WinoGrande 5-shot~\cite{sakaguchi2019winogrande} (\textbf{Wino}) and GSM8K 5-shot~\cite{cobbe2021training} (\textbf{GSM}).
Also, as Stable LM 2 is a general-purpose foundational model, we further assess natural language understanding capabilities by evaluating English and machine-translated versions of LAMBADA. 
All evaluations are performed with the Language Model Evaluation Harness framework~\cite{eval-harness}\footnote{\url{https://github.com/Stability-AI/lm-evaluation-harness/tree/stablelm-2/multilingual-bench}}. 
As shown in Table~\ref{tab:pre-training-eval}, Stable LM 2 1.6B (\textit{stablelm-2-1-6b}) outperforms other base models by a significant margin.
Similarly, the instruction-tuned version (\textit{stablelm-2-1-6b-dpo}) improves on Microsoft's Phi-1.5 by two average points while lagging behind the larger Phi-2.0 on few-shot accuracy.
Performance versus Google's Gemma 2B (2.5B parameters) is also remarkable.

\begin{table}[t]
\centering
\begin{tabular}{G G G}
\toprule
\multicolumn{1}{l}{\textbf{Model}} & \textbf{Size} & \textbf{MT-Bench} \\
\cmidrule(lr){1-1}
\cmidrule(lr){2-3}
%\textbf{Model} & \textbf{Size} & \textbf{MT-Bench} \\ \hline
\multicolumn{1}{l}{\textbf{Mistral-7B-Instruct-v0.2}} & 7B & 7.61 \\
\multicolumn{1}{l}{\textbf{Llama2-Chat}} & 70B & 6.86 \\
\multicolumn{1}{l}{\textbf{stablelm-zephyr-3b}} & 3B & 6.64 \\
\multicolumn{1}{l}{\textbf{MPT-30B-Chat}} & 30B & 6.39 \\
\rowcolor{PaleBlue} \multicolumn{1}{l}{\textbf{stablelm-2-zephyr-1.6b}} & 1.6B & 5.42 \\
\multicolumn{1}{l}{\textbf{Qwen-1.5-1.8B-Chat}} & 1.8B & 5.29 \\
\multicolumn{1}{l}{\textbf{gemma-2b-it}} & 2.5B & 5.19 \\
\multicolumn{1}{l}{\textbf{Falcon-40B-Instruct}} & 40B & 5.17 \\
\multicolumn{1}{l}{\textbf{dolphin-2.6-phi-2}} & 2.7B & 4.93 \\
\multicolumn{1}{l}{\textbf{phi-2}} & 2.7B & 4.29 \\
\multicolumn{1}{l}{\textbf{TinyLlama-1.1B-Chat-v1.0}} & 1.1B & 3.46 \\
\midrule
\end{tabular}
\caption{MT-Bench results}
\label{tab:mt-bench}
\end{table}

\subsection{Multilingual Evaluations}
We assess knowledge and reasoning in the multilingual setting for non-English languages seen during pre-training by evaluating on ChatGPT-translated versions of ARC, HS, TQA, and MMLU (\cite{lai2023okapi}). In addition, we test next-word prediction capabilities using the machine-translated LAMBADA datasets from \cite{eval-harness}.
After manual inspection by native speakers, we have deemed existing machine translations \footnote{\url{https://huggingface.co/datasets/EleutherAI/lambada_openai}} too noisy to draw accurate performance signals from. We instead evaluate multilingual next-word prediction with new translations which are made available for researchers \footnote{\url{https://huggingface.co/datasets/marcob/lambada_multilingual}}.

The zero-shot results are presented in Tab.~\ref{tab:pre-training-eval-multilingual} and highlight Stable LM 2's superior performance compared to models even twice its size.

% We assess knowledge and reasoning in the multilingual setting for non-English languages seen during pre-training by evaluating on ChatGPT-translated versions of ARC, HS, TQA, and MMLU (\cite{lai2023okapi}). In addition, we test next-word prediction capabilities in French, German, Italian, and Spanish using the machine-translated LAMBADA datasets from \cite{eval-harness}. After manual inspection by native speakers, we've deemed these machine translations \footnote{\url{https://huggingface.co/datasets/EleutherAI/lambada_openai}} too noisy to draw accurate performance signals from and advise future researchers to adopt alternative multilingual next-word prediction benchmarks.
% The zero-shot results are presented in Tab.~\ref{tab:pre-training-eval-multilingual} and highlight Stable LM 2's superior performance compared to models even twice its size.

\subsection{MT Benchmark Evaluations}

Finally, we also test the conversational skills of our model on the popular multi-turn benchmark \textbf{MT-Bench}~\cite{zheng2023judging}. The results are provided in Fig.~\ref{fig:MT-bench} and Tab.~\ref{tab:mt-bench}. While lagging behind much more powerful models such as Mistral 7B Instruct v0.2 (more than 4x the size of Stable LM 2), our model delivers better chat performance and beats both Phi-2, Gemma 2B and TinyLLaMA 1.1B by a wide margin despite the larger size of the former.

% \begin{tabular}{Q M M M M M M M M}\toprule
% \multicolumn{1}{ l }{\textbf{Model}} & \textbf{All}& \textbf{EN} & \textbf{DE} & \textbf{ES} & \textbf{FR} & \textbf{IT} & \textbf{NL}$^\ddagger$ & \textbf{PT}$^\ddagger$\\[0.1ex]
% \cmidrule(lr){1-1}
% \cmidrule(lr){2-9}
% \rowcolor{PaleBlue} \multicolumn{1}{ l }{\textbf{stablelm-2-zephyr-1\_6b}\dag} &40.0& 	49.6& 	37.0& 	36.2& 	39.1& 	39.0& 	39.5& 	39.7\\ [0.1ex]

\section{Inference and Quantization}\label{sec:inference}
This model represents a substantial leap towards making advanced generation capabilities available directly on-device without the computational overhead of larger models. We believe this model strikes a great balance between remarkable efficiency and effectiveness in inference tasks when paired with inference frameworks and quantization methods. As part of our release, we provide quantized weights of \textit{stablelm-2-1-6b} supported on popular inference libraries such as llama.cpp \footnote{\url{https://github.com/ggerganov/llama.cpp}}, Apple MLX~\cite{mlx2023} and Intel OpenVINO \footnote{\url{https://github.com/openvinotoolkit/openvino}}.

\subsection{Quantization}

We make available quantization files for various models and formats to support easier
integration with different inference frameworks, including:
\begin{itemize}
\item Two 4-bit quantized models: Q4\_0, Q4\_1 and a 5-bit quantized model: Q5\_K\_M GGUF
\item INT4 for OpenVINO quantized with Intel's Neural Network Compression Framework (NNCF)
\item INT4 for MLX quantized with MLX
\end{itemize}
%\carlos{Maybe explain a bit what these different models are?}

% \carlos{Seems INT4 for OpenVINO is missing from Table \ref{tab:quant-eval}?}

% \textcolor{red}{ashish: We do not have an eval harness to run it on so I don't have numbers for that. We should omit from the table}
%
These quantization files can be found in the model's
\href{https://huggingface.co/stabilityai/stablelm-2-zephyr-1_6b/tree/main}{Hugging Face repository} for the
convenience of developers and researchers working with our models. We aim to facilitate smoother deployment experiences across various deep-learning framework
ecosystems by offering a range of quantization formats.

% \carlos{Shouldn't we show the amount of performance degradation for those quantized models?}

\subsection{Throughput}
%We provide in Tab.\ref{tab:quant-config} an overview of the system environments utilized in our experimental evaluations.  \marco{maybe move to acknowledgements} We acknowledge support from the MLX team's in facilitating the data collection process through their MLX framework.
%\textcolor{red}{ashish: Leaving placeholder for Intel OpenVINO if we have time}\\
% \begin{table}
% \centering
% \noindent
% \begin{tabular}{c c c}
%   \toprule
%   \textbf{Device} & \textbf{CPU} & \textbf{Memory} \\
%   \cmidrule(lr){1-3}
%   MacBook Pro & Apple M2 Pro Max & 16GB \\[0.1ex]
%   Laptop (x86) & Intel Core i7 & 16GB \\[0.1ex]
%   Mac Mini (M2) & Apple M2 & 8GB \\[0.1ex]
%   \cmidrule(lr){1-3}
% \end{tabular}
%   \caption{Experimental system configurations}
%   \label{tab:quant-config}
% \end{table}
In Tab.~\ref{tab:quant-tput} we provide throughput numbers obtained from our model running on consumer-grade devices and the system environments utilized. 
Our initial runs showcase that when using a lower precision, we are able to achieve almost 2x performance in throughput. 
Note that these figures are provided as a reference, and they are not the result of rigorous benchmarking but are rather intended to give users a practical insight into what they can expect in terms of performance on commonly used devices. Likewise, as lower precision quantization is expected to reduce the model's performance, we encourage researchers and developers to assess the potential degradation in real-world scenarios.

% Our initial runs showcased in table \ref{tab:quant-tput} that when using a lower precision we are able to achieve almost 2x performance in throughput. 
% We encourage researchers and developers to build upon this foundation by testing and optimizing our models on an even wider range of devices.
%
\begin{table}
\centering
\noindent
\begin{tabular}{G G N G H }\toprule
\multicolumn{1}{ l }{\textbf{Framework}} & \textbf{CPU}  & \textbf{Precision} & \textbf{Throughput (Tok/s)} & \textbf{Power consumption (W)}\\[0.1ex]
\cmidrule(lr){1-1}
\cmidrule(lr){2-5}
\multicolumn{1}{ l }{MLX} & M2& FP16 & 71 & 6\\ [0.1ex]
\multicolumn{1}{ l }{Mac Mini}& (8GB)& INT4 & 127 & 11\\ [0.1ex]
\cmidrule(lr){1-1}
\cmidrule(lr){2-5}
% \multicolumn{1}{ l }{Intel OpenVINO} &Intel Core i7 &FP16 &-&-\\ [0.1ex]
% \multicolumn{1}{ l }{x86 Laptop} & (16GB)& INT4 & -& -\\ [0.1ex]
% \cmidrule(lr){1-1}
% \cmidrule(lr){2-5}
\multicolumn{1}{ l }{GGUF} &M2 Pro Max &FP16 &46 & 14\\ [0.1ex]
\multicolumn{1}{ l }{2023 MacBook Pro} & (16GB) &INT4 & 99& 14 \\ [0.1ex]
\midrule
\end{tabular}
\caption{Throughput and power usage on various devices using different quantization frameworks. We employ a batch size of 1 for all benchmarks. INT4 numbers for GGUF were collected using Q4\_0 quantization. }
\label{tab:quant-tput}
\end{table}

\section{Future Work} \label{sec:future}
There is a number of research avenues we would like to explore to further improve the model:
\begin{enumerate}
  \item \textbf{Data}. In this work, we focused on publicly available data. In particular, most of the data comes from web-crawled content, as is common for most models. This data is known to contain many low-quality documents~\cite{elazar2023whats} that can potentially harm training.
  We believe there is significant potential in smart filtering, re-writing, and synthetic data generation with strong models.
  \item \textbf{Hallucination Mitigation}.
  Language models are prone to generating incorrect or misleading information, and small language models are even more prone to doing so. Finding reliable ways to detect hallucinations in these models will unlock new use cases in areas that are sensitive to hallucinations.
  \item \textbf{Long Contexts and Retrieval}. 
  The ability to retrieve information across long context windows is essential for applications such as chat models or dataset integration. Accordingly, in Appendix~\ref{sec:context}, we explore the current capabilities and limitations of StableLM2 1.6B on the Needle-in-the-Haystack task. Going forward we plan to further build upon this work as well as to extend our models to context lengths beyond 4k.
  \item \textbf{Conditional Computation}. Small models are often capacity-constrained -- that is, with the current training approaches, they lack the capacity to process and exploit all of the training data. Recently, ideas such as Mixture of Experts have been successfully applied to take a dense model and extend it to contain more parameters that are selectively applied to certain inputs (for instance, via sparse upcycling~\cite{komatsuzaki2022sparse}).
  Importantly, if each token selects only one expert, the overall inference FLOPs do not significantly change.
  Applying this to the Stable LM 2 1.6B model is a natural extension we will investigate.
\end{enumerate}

\section{Environmental and Societal Impact} \label{sec:impact}
\subsection{Carbon Footprint}
The training of Stable LM 2 has consumed energy with associated carbon dioxide emissions. In line with \cite{touvron2023llama} we report our carbon footprint based on the formula
\begin{align*}
    \text{Total Wh} = \text{GPU-h} \times \text{power consumption} \times \text{PUE}
\end{align*}
where the Power Usage Effectiveness is set to 1.1. We trained Stable LM 2 for $\approx 92,000$ GPU-hours, giving a total power consumption of 30MWh considering our average power usage. The tons of emitted carbon tCO$_2$eq can be estimated using the US national average carbon intensity factor of 0.385 kg CO$_2$eq/KWh, leading to a final figure of 11 tCO$_2$eq.

\subsection{Societal impact}
Stability AI is committed to releasing open models to help improve access to foundational AI technology. Open access to model weights enables researchers to inspect models for suitability and vulnerabilities, test the effectiveness of different optimization strategies, and correct for biases observed in the model. To that end, this model is released under an open non-commercial license. However, open release can introduce challenges in assessing the societal impact of a model. For example, Stability AI does not have direct visibility into downstream applications of Stable LM 2 1.6B, the distribution of applications by sector, or the distribution of model usage by geography. Since the model is released with a noncommercial license, we expect a limited number of applications outside of fine-tuning or evaluation interfaces and a limited number of third parties affected by the model.

% This data, if available at all, is held by intermediary distribution platforms such as GitHub and Hugging Face repositories. As a proxy for direct impact, we monitor monthly and aggregate download figures for Stable LM 2 1.6B through Hugging Face. 
% In the first three weeks since its release on January 19, we have observed 18,685 downloads of the base model and 10,168 downloads of the instruction fine-tuned model, for a total of 28,853 downloads. 
We will continue to monitor openly released fine-tuned models to understand the extent of fine-tuning research or development activity that uses Stable LM 2 1.6B as a base model, including the evaluation results of these derivative models.

\section{Conclusion} \label{sec:conclusion}
In this report, we introduced Stable LM 2 1.6B, a compact decoder-only language model trained on multilingual datasets. It fluidly handles up to seven languages: English, Spanish, German, Italian, French, Portuguese, and Dutch.
To ensure that the community can reproduce our run, we detail all datasets used during training --with the \emph{exact} data mix-- and our newly designed learning rate schedule.
We also conduct extensive model evaluations and comparisons with other similarly-sized models, demonstrating Stable LM 2 1.6B's exceptional performance.
Finally, we profile the model on common edge computing architectures.
We hope the current report contributes to the improvement and further research on small language models.

\section*{Acknowledgments}
We thank our awesome MLOps team members, particularly Richard Vencu, for the support provided.
We also thank Christian Laforte, Cedric Wagrez, and Jerry Chi for their feedback, useful ideas, and comments.

\newpage
{\small
\bibliographystyle{plain}
\bibliography{literature}
}
%%%%%%%%%%%%%%%%%%%%%%%%%%%%%%%%%%%%%%%%%%%%%%%%%%%%%%%%%%%%%%%%%%%%%%%%%%%%%%%
%%%%%%%%%%%%%%%%%%%%%%%%%%%%%%%%%%%%%%%%%%%%%%%%%%%%%%%%%%%%%%%%%%%%%%%%%%%%%%%
% APPENDIX
%%%%%%%%%%%%%%%%%%%%%%%%%%%%%%%%%%%%%%%%%%%%%%%%%%%%%%%%%%%%%%%%%%%%%%%%%%%%%%%
%%%%%%%%%%%%%%%%%%%%%%%%%%%%%%%%%%%%%%%%%%%%%%%%%%%%%%%%%%%%%%%%%%%%%%%%%%%%%%%
\newpage
\appendix
\include{appendix}
\section{Data Ablations} \label{sec:data-ablations}
How to select the optimal training mix for pre-training from a set of sources is an open problem. Tuning weights based on downstream tasks~\cite{chowdhery2022palm, du2022glam} can be extremely costly and bears the risk of overfitting on particular tasks as well as exploiting data leakage. While the computationally cheap, principled approach introduced in~\cite{xie2023doremi} is promising, we found it delivers sub-optimal weights when the data sources are highly imbalanced and have different information content (e.g., large web sources vs curated datasets).
Furthermore, multilingual evaluations introduce a more explicit dependence on the tokenizer, with increased noise due to the lack of high-quality, non-machine-translated benchmarks.
We, therefore, aim to find general guiding principles that are expected to hold against changes in the tokenizer or in the absence of high-quality benchmarks for each data category while keeping the cost of these ablations low.

We trained a set of 1B models on a total of 100B tokens sampled according to Tab.~\ref{tab:data-ablations}.
\begin{table}
\centering
\noindent
\begin{tabular}{N >{\columncolor{PaleBlue}} N  >{\columncolor{PaleGreen}} N >{\columncolor{PaleGreen}} N >{\columncolor{PaleGreen}} N >{\columncolor{Pink}} N >{\columncolor[gray]{0.8}} N }\toprule
\multicolumn{7}{c }{\textbf{Sampling weights}} \\ [1ex] 
\multicolumn{1}{ c }{\textbf{Source}} & \textbf{Control} & \textbf{Mix 1}
& \textbf{Mix 2}	&  \textbf{Mix 3} & \textbf{Mix 4} & \textbf{Mix 5} \\ [1ex]
\cmidrule(lr){1-1}
\cmidrule(lr){2-7}
\multicolumn{1}{ l }{Cultura-x En} & 0.6894 & 0.5694	& 0.3294	&  0.1494& 0.49  & 0.49 \\ [1ex]
\multicolumn{1}{ l }{Cultura-x De} & 0. & 0.03 & 0.06 & 0.09 & 0. & 0. \\ [1ex] 
\multicolumn{1}{ l }{Cultura-x Es} & 0. & 0.03 & 0.06 & 0.09 & 0.  & 0.\\ [1ex] 
\multicolumn{1}{ l }{Cultura-x Fr} & 0. & 0.03 & 0.06 & 0.09  & 0.  & 0.\\ [1ex] 
\multicolumn{1}{ l }{Cultura-x It} & 0. & 0.03 & 0.06 & 0.09  & 0.  & 0.\\ [1ex] 
\multicolumn{1}{ l }{Cultura-x Pt} & 0. & 0.03 & 0.06 & 0.09  & 0.  & 0.\\ [1ex] 
\multicolumn{1}{ l }{Cultura-x Nl} & 0. & 0.03 &  0.06 & 0.09  & 0.  & 0.\\ [1ex] 
\multicolumn{1}{ l }{Starcoder} & 0.0702 & 0.0702 & 0.0702 & 0.0702 & 0.0497 & 0.28 \\ [1ex] 
\multicolumn{1}{ l }{Others} & 0.2292 & 0.2292 & 0.2292 & 0.2292 & 0.4648 & 0.2292 \\ [1ex] 
\midrule
\end{tabular}
\caption{
Data ablations with corresponding sampling weights. Column 2 (\textbf{Control}) is our reference, containing only English and code data, with a standard, significant amount of web data. \textbf{Mix 1-3} test our strategy for adding multilingual data, capping non-English sources to a fixed percentage. In \textbf{Mix 4-5} we reduce the amount of web data, increasing respectively books and academic sources and code. 
Source \textbf{Others} contains the same data as in Tab.~\ref{tab:data-mix} from the categories: Academic, Books, Social, Law, Math, and Wiki. In \textbf{Mix 4}, we uniformly upsample Academic and Books sources.
}
\label{tab:data-ablations}
\end{table}
Evaluations of each model on English and non-English benchmarks are shown in Tab.~\ref{tab:data-evals}.
We observe the following trends
\begin{itemize}
    \item Contrary to~\cite{muennighoff2023scaling}, we find less conclusive evidence that code can be used as a neutral filler for training data as increasing the amount of code leads to a degradation in language model performances. We leave a more thorough exploration of this to future work, including math and reasoning tasks that may benefit from a higher proportion of code data.
    \item Performance on non-English benchmarks increases for each language by adding any amount of data in that same language. However, this increase saturates very fast and we observe only modest gains beyond $6\%$.
    We hypothesise that this might be due to the lack of high-quality, structured data sources in non-English languages, which we only sample from the web.
    \item Upsampling Academic and Books sources improves downstream performance over the control run, particularly in natural language understanding.
\end{itemize}
\begin{table}
\centering
\noindent
\begin{tabular}{P P G P P}\toprule
%\begin{tabular}{N N N N N }\toprule
\multicolumn{1}{ c }{\textbf{Data}} & \textbf{Avg} & \textbf{English Commonsense} & \textbf{LAMBADA}
& \textbf{Okapi}	\\ [1ex]
%\multicolumn{1}{ N }{\textbf{Source}} &\textbf{CO (\%)} & \textbf{HC (ppm)} & \textbf{Nox (ppm)} & \textbf{$\boldsymbol{\mathrm{CO_2}}$ (\%)} & \textbf{CO (\%)} & \textbf{HC (ppm)} & \textbf{Nox (ppm)} & \textbf{$\boldsymbol{\mathrm{CO_2}}$ (\%)} \\ 
\cmidrule(lr){1-1}
\cmidrule(lr){2-5}
\multicolumn{1}{ c }{\textbf{Control}} & 38.76 & \textbf{66.51} & 30.93 & 33.98 \\ [1ex]
\multicolumn{1}{ c }{\textbf{Mix 1}} &39.91&	63.92&	34.25&	35.39 \\ [1ex] 
\multicolumn{1}{ c }{\textbf{Mix 2}} &\textbf{40.69}&	64.71&	\textbf{35.74}&	\textbf{35.93} \\ [1ex] 
\multicolumn{1}{ c }{\textbf{Mix 3}} &39.87&	63.02&	35.22&	35.25 \\ [1ex] 
\multicolumn{1}{ c }{\textbf{Mix 4}} & 39.41 & 65.87 & 32.73 & 34.58 \\ [1ex] 
\multicolumn{1}{ c }{\textbf{Mix 5}} & 38.66 & 64.95 & 31.4 & 34.07 \\ [1ex] 
\midrule
\end{tabular}
\caption{
Downstream evaluations of the models considered in our data ablations. English commonsense includes: \textbf{ARC-Easy}, \textbf{PIQA}, \textbf{SciQ}, \textbf{WinoGrande}. Each score is averaged over EN, DE, ES, FR, and IT. }
\label{tab:data-evals}
\end{table}
%
% \begin{figure}
%     \centering
%     \includegraphics[width=12cm]{figs/multilingual-evals.pdf}
%     \caption{Multilingual evals. \marco{Same info as the table, let's see if it makes sense to keep}}
%     \label{fig:multilingual-evals}
% \end{figure}
%

\section{Scheduler Ablations}\label{app:scheduler}
\begin{figure}
    \centering
    \includegraphics[width=\linewidth]{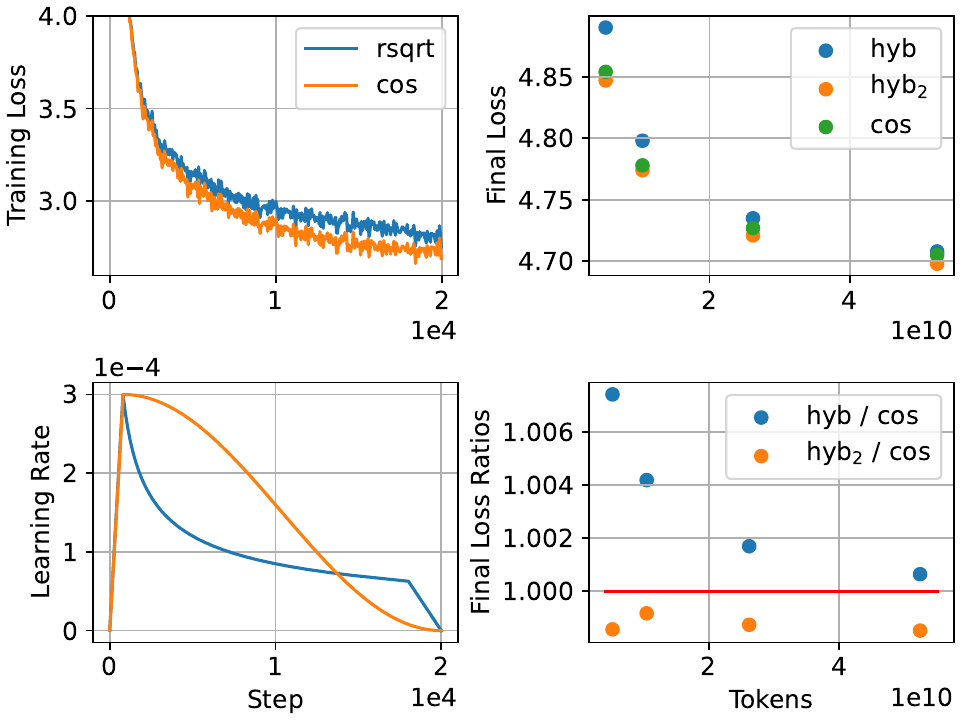}
    \caption{Learning rate scheduler ablations. Left figures: comparison between rsqrt and cosine decay scheduler. Right figures: final loss of models trained with the two variants of the hybrid scheduler and cosine for different number of training tokens (top). The difference in final loss between the 3 schedulers is $< 1\%$ (bottom).}
    \label{fig:schedulers-comparison}
\end{figure}

\cite{hoffmann2022training} shows that the widely adopted cosine learning rate decay achieves optimal performance only when performing a full cosine period, forcing practitioners to fix the number of steps beforehand. As multi-epoch training performs well for LLMs~\cite{muennighoff2023scaling, luukkonen2023fingpt, StableLM-3B-4E1T}, and larger and cleaner data sources are made accessible by the OS community, it becomes more and more important to alleviate this limitation. 
To this end, we experimented with the "inverse square root" (rsqrt) learning rate scheduler~\cite{raffel2023exploring} Eq.~\ref{eq:inv-sqrt}
\begin{align}\label{eq:inv-sqrt}
    \frac{1}{\sqrt{\text{max}\left(i, k\right)}} 
\end{align}
where $i$ is the current iteration and $k$ the number of warmup steps. As the scheduler is strictly convex and reaches zero asymptotically, it can be used to train for infinite iterations.
However, in standard scenarios, of which we show an example in Fig.~\ref{fig:schedulers-comparison}, rsqrt consistently underperforms cosine. We make the comparison by decaying the learning rate to 0 in both cases, with a linear cool down for the last $10\%$ of the steps for the rsqrt scheduler.

A sharp difference between the two schedulers is how they start from the peak learning rate: with a flat slope, negative second derivative for cosine and a large slope, positive second derivative for rsqrt. This allows rsqrt to escape the high learning rate region quickly, which we identified as the main cause of the performance gap.
We then get the best of both worlds by combining both schedulers into a single function that is again the cosine for the first section, and smoothly switches to rsqrt, as described in Eq.~\ref{eq:scheduler}.

Finally, we experimented with two different versions of our scheduler, whose performance we show in the right panels of Fig.~\ref{fig:schedulers-comparison}: hybrid (\textbf{hyb}) corresponds to our final scheduler defined in Eq.~\ref{eq:scheduler}, while for hybrid$_2$ (\textbf{hyb$_2$}) we double the cosine period and move the turning point at half the training steps instead of at a fourth. We attribute the difference in performance between the two versions to the different average learning rates. For the set of experiments in Fig.~\ref{fig:schedulers-comparison}, integrating the scheduler over the training volume, we obtain an average learning rate of $1.5e-4$, $1.13e-4$ and $1.67e-4$ for cosine, hyb and hyb$_2$ respectively. We leave to future work a proof that schedulers with the same average value lead to statistically equivalent models under mild conditions, such as monotonicity and fixed endpoints.
%
% \begin{figure}
%     \centering
%     \includegraphics[width=12cm]{figs/hybrid-scheduler-evals.pdf}  
%     \caption{\marco{TODO: finalize names, possibly remove hybrid 1 and rewrite better.}\\
%     Comparison between Cosine and Hybrid schedulers. Up: final training loss smoothed with a Gaussian filter. Bottom: ratio of losses shows how the difference is minimal}
%     \label{fig:hybrid-schedulers-evals}
% \end{figure}
%
\section{Evaluation of Performance Across Different Sized Context Windows}\label{sec:context}

The Needle-in-a-haystack test, as introduced in~\cite{needle_eval}, is commonly used to assess the retrieval capabilities of LLMs across different context window sizes. Following the methodology of~\cite{needle_eval}, a fact ("the needle") is embedded within a context of unrelated essays ("the haystack"). 
The model under evaluation is tasked to answer a question requiring the retrieval of the needle from the context. 
The evaluation is carried out systematically by placing the needle at 35 different depths within the context and comparing the results across context window sizes from 500 to 4000 tokens. Notably, these window sizes are specifically chosen to assess the performance within our trained context size of 4096 tokens. An AI judge, typically GPT-4~\cite{openai2023gpt4}, scores the answers from 1 to 10 based on wether or not the fact was correctly retrieved.

\begin{figure}
    \centering
    \includegraphics[width=\linewidth]{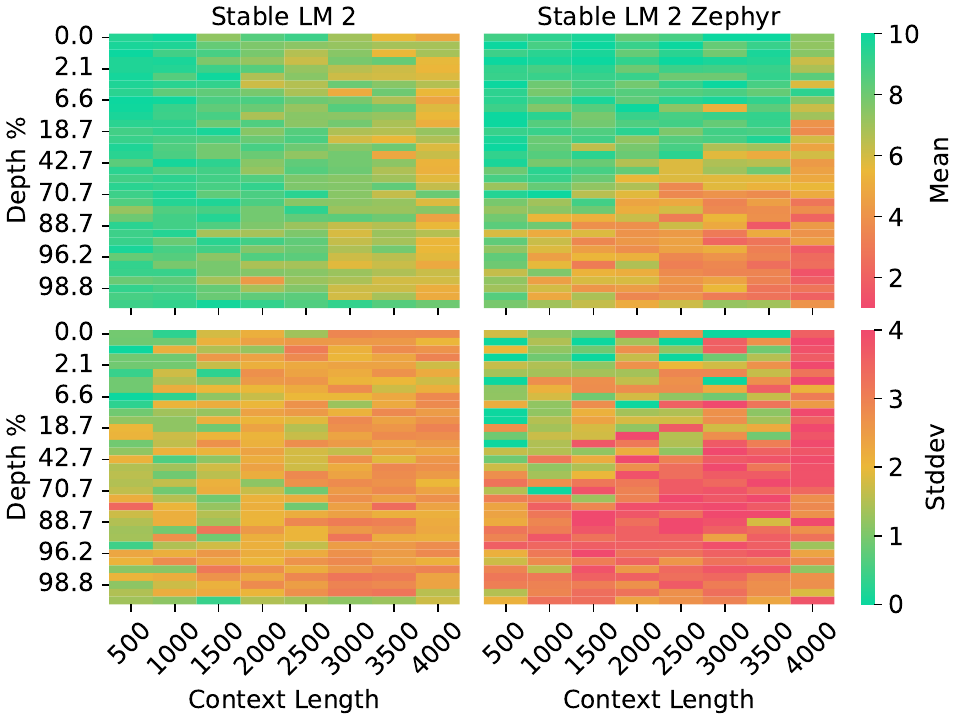}
    \caption{Needle-in-a-haystack evaluation of Stable LM 2 on context window sizes from 500 to 4000 tokens.}
    \label{fig:needle_eval}
\end{figure}

While performing the evaluation, we noticed that the order of the context was not fixed or controlled by a seed, leading in some cases to significantly different scores. For this reason, we show in Fig.~\ref{fig:needle_eval} the average results over 10 different runs, together with the corresponding standard deviation. For Stable LM 2, we employ the prompt of~\cite{fu2024data}, whereas for our fine-tuned version we use the official repository as is. 
Averaging the scores over the evaluated grid, we observe a slight degradation from $\approx7.7$ for our base model to $\approx6.8$ for the fine-tuned version. The different prompt structures make it hard to directly compare the results, however, we will investigate in future work how different elements such as the distribution of the document lengths and the attention mask correlate with this behavior.

% \begin{figure}
%     \centering
%     \includegraphics[width=9cm]{figs/doc-lens.pdf}
%     \caption{Distribution of document lengths. Average document length is $\approx$ 842, with 50$\%$ of the documents being shorter than 380 tokens.}
%     \label{fig:doc_lens}
% \end{figure}
% 
\begin{figure}
    \centering
    \includegraphics[width=9cm]{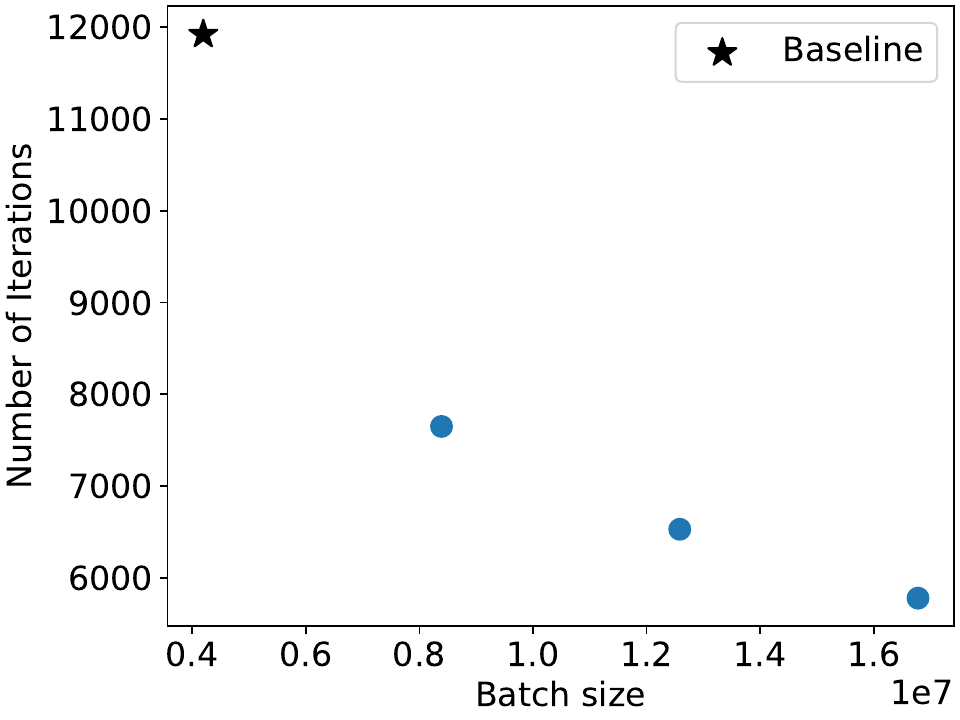}
    \caption{Number of iterations performed at various batch sizes to match the same final loss.}
    \label{fig:global-batch}
\end{figure}

\section{Global Batch Size}\label{sec:batch}

The role of the batch size in stochastic optimization convergence is illustrated in \cite{mccandlish2018empirical} as follows.  
The loss $L({\theta})$ of a model parameterized by $\theta$ over a training set $D$ is estimated by independently drawing random samples from $D$ to form a batch $B$.
Thus, the loss gradient used to update the model's parameters is given by
\begin{align}
    \nabla L\left({\theta}\right) = \frac{1}{|B|} \sum_{i=1}^{|B|} \nabla_{\theta} L_{x_i} \left(\theta\right)
\end{align}
and its variance scales with $1 / |B|$. In other words, a bigger batch has a lower variance and thus provides a more accurate estimate of the gradient.
A more accurate gradient suggests that we should increase the learning rate accordingly to converge faster; however, in practice, this is far from trivial and requires special handling such as fine-tuning the learning rate scheduler or performing per-layer updates~\cite{goyal2018accurate, you2020large}. 

To choose the global batch size for Stable LM 2, we make the following assumptions:
\begin{enumerate}
    \item We do not change the learning rate based on the batch size
    \item The batch size can be approximately scaled at no cost.
\end{enumerate}
With $1.$, we are, in principle, giving up on potential further gains by tuning the step size to the gradient noise. In practice, we notice that using a large learning rate at the boundary between convergence and divergence of our ablations is more than enough to compensate for this.
$2.$ follows from a combination of theoretical results, hardware optimizations, and increased training data availability. \cite{muennighoff2023scaling} empirically demonstrated how multiple epochs of repeated data are as good as fresh data in LLMs pre-training, while the computational overhead of increasing data parallel workers is minimal.

The data volume available for pre-training is consistently increasing through larger datasets, and there are promising results of multi-epoch training. Hence, 
we explore training with larger batch sizes, which requires more overall training tokens to reach the final loss but significantly decreases training time. 
To determine our final batch size, we start by training a baseline with a batch $B_0$ of 4M tokens on $T_0 = 50B$ total training tokens. Subsequently, we train new models from scratch with batches $B_i = $ 8M, 12M, and 16M, increasing $T_i$ tokens until the same final loss is reached. 
We do this with an rsqrt scheduler as the number of required training steps is unknown beforehand. 

In Fig.~\ref{fig:global-batch}, we show the number of iterations $B_i / T_i$ which are required to match the baseline loss. Assuming for simplicity that the iteration time is independent of the batch size, this gives an upper bound on the speed up we can achieve by increasing the batch.

In the regime considered, we observe that an increase in batch size leads to a decrease in the number of iterations required to match the baseline loss all the way to the biggest batch considered of $16.7$M tokens, which speeds up training by a factor 2x. However, to achieve an equal loss, we require $T_i = 96B$ training tokens, which is a factor 1.96 increase compared to the baseline.
Therefore to train Stable LM 2, we opt for a global batch of $8,388,608$, which with the layout of Tab.~\ref{tab:training-config} offers the best compromise between decrease in training time and additional required training tokens.
% 
% \begin{figure}
%     \centering
%     \includegraphics[width=9cm]{figs/batch.pdf}
%     \caption{Number of iterations performed at various batch sizes to match the same final loss.}
%     \label{fig:global-batch}
% \end{figure}
%

\newpage
\section{Restructed Pre-training Sources}

While many sources are restructured in \cite{yuan2022restructured}, in this work, we have only considered those listed in Tab.\ref{tab:restruct-data} with commercially viable licenses such as BIGPATENT, CLOTH,  SciTLDR, TriviaQA, WordNet, WikiHow, etc. We also restructure additional sources with similar licensing compatibility and list them below.

\begin{table}[ht]
\centering
\begin{tabular}{ll}
\toprule
\textbf{Dataset} & \textbf{Prefix} \\
\cmidrule(lr){1-1}
\cmidrule(lr){2-2}
Banking77 \cite{Casanueva2020} & banking77 \\
BigPatent \cite{DBLP:journals/corr/abs-1906-03741} & big\_patent \\
BIOSSES \cite{souganciouglu2017biosses} & biosses \\
BLBooksGenre \cite{british_library_genre} & TheBritishLibrary/blbooksgenre \\
CodeComplex \cite{JeonBHHK22} & codeparrot/codecomplex \\
CoEdIT \cite{raheja2023coedit} & grammarly/coedit \\
CLOTH \cite{DBLP:journals/corr/abs-1711-03225} & AndyChiang/cloth \\
CommonGen \cite{lin-etal-2020-commongen} & common\_gen \\
FigQA \cite{https://doi.org/10.48550/arxiv.2204.12632} & nightingal3/fig-qa \\
FeasibilityQA \cite{gupta-etal-2023-john} & jon-tow/feasibility\_qa \\
Flan 2021 \cite{longpre2023data} & DataProvenanceInitiative/flan2021\_submix\_original \\
Flan Chain of Thought \cite{longpre2023data} & DataProvenanceInitiative/cot\_submix\_original \\
Flan NIv2 \cite{longpre2023data} & DataProvenanceInitiative/niv2\_submix\_original \\
Flan T0 \cite{longpre2023data} & DataProvenanceInitiative/t0\_submix\_original \\
HelpSteer \cite{wang2023helpsteer} & nvidia/HelpSteer \\
IMDB Review \cite{maas-EtAl:2011:ACL-HLT2011} & ajaykarthick/imdb-movie-reviews \\
Joke Explanation \cite{longpre2023data} & dim/joke\_explaination \\
MBPP \cite{austin2021program} & mbpp \\
NarrativeQA \cite{kocisky-etal-2018-narrativeqa} & narrativeqa \\
PuzzleQA \cite{zhao2023solving} & Jingmiao/PUZZLEQA \\
SciTLDR \cite{cachola2020tldr} & allenai/scitldr \\
Self-Instruct Starcoder \cite{self-instruct-starcoder} & codeparrot/self-instruct-starcoder \\
SQL Create Context \cite{b-mc2_2023_sql-create-context} & b-mc2/sql-create-context \\
StepGame \cite{stepGame2022shi} & tasksource/stepgame\\
TRACIE \cite{ZRNKSR21} & tasksource/tracie \\
TriviaQA \cite{2017arXivtriviaqa} & trivia\_qa \\
WikiHow & wikihow \\
WordNet \cite{mccrae-etal-2019-english} & jon-tow/open-english-wordnet-synset-2023 \\
Yahoo Answers Topics & yahoo\_answers\_topics \\
\midrule
\end{tabular}
\caption{The original sources for restructured and instruction pre-training datasets can be found at \url{https://huggingface.co/datasets/} followed by the provided prefix.}
\label{tab:restruct-data}
\end{table}
\end{document}